\documentclass[]{spie}  

\usepackage{amsmath,amsfonts,amssymb}
\usepackage{graphicx}
\usepackage[colorlinks=true, allcolors=blue]{hyperref}

\title{Probabilistic Machine Learning for Short-Term Atmospheric Seeing Forecasting}

\author[a]{Mary Joe Medlej}
\author[b,c]{Rahul Srinivasan}
\author[a]{Simon Prunet}
\author[a]{Aziz Ziad}
\author[a]{Christophe Giordano}
\affil[a]{Université Côte d’Azur, Observatoire de la Côte d’Azur, J.L. Lagrange, Parc Valrose 06108 Nice
Cedex 2, France}
\affil[b]{SISSA, Via Bonomea 265, 34136 Trieste, Italy}
\affil[c]{IFPU - Institute for Fundamental Physics of the Universe, Via Beirut 2, 34014 Trieste, Italy}

\authorinfo{Further author information: \\mary-joe.medlej@oca.eu\\rsriniva@sissa.it}

\begin{document} 
\maketitle

\begin{abstract}
Accurate short-term forecasting of atmospheric seeing is of significant interest for astronomical observations and free-space optical communication systems, where rapid variations in optical turbulence can affect system performance and operational planning. In this work, we benchmark machine learning models for predicting atmospheric seeing up to two hours in advance, with a particular focus on probabilistic forecasting and uncertainty quantification. We compare two probabilistic approaches, Gaussian Processes (GPs) and a normalizing-flow-based model (\textsc{FloTS}), against deterministic recurrent neural network (RNN) and long short-term memory (LSTM) baselines. All models are trained and evaluated on seeing measurements from the Maunakea summit. We assess the models using both deterministic metrics, including the root mean square error and the Pearson correlation coefficient, and probabilistic metrics that evaluate the quality and calibration of predictive uncertainty estimates. The results show that deep learning approaches outperform standard recurrent networks for short-term forecasting, with LSTM and \textsc{FloTS} achieving the best overall predictive accuracy. While Gaussian Processes provide a useful probabilistic baseline, their predictive distributions are limited by the Gaussian assumption. \textsc{FloTS} combines competitive forecasting accuracy with flexible uncertainty quantification and produces the most informative probabilistic forecasts.
\end{abstract}

\section{INTRODUCTION}
\label{sec:intro}  
Atmospheric optical turbulence (OT) arises from fluctuations in the refractive index of air caused by variations in temperature, pressure, and wind. These fluctuations distort the propagation of optical waves through the atmosphere and constitute a major limitation for ground-based optical systems. In astronomy, turbulence degrades image quality and limits the angular resolution achievable by telescopes, an effect commonly quantified by the atmospheric seeing parameter. In free-space optical (FSO) communications, turbulence induces signal fluctuations and beam distortions that can reduce link reliability and communication performance.

To mitigate the effects of turbulence, adaptive optics (AO) systems are widely used in both astronomical and optical communication applications. By measuring and correcting wavefront distortions in real time, AO systems can substantially improve image quality and signal transmission. However, these systems remain fundamentally reactive, relying on measurements of the current atmospheric state. Rapid changes in turbulence conditions can therefore limit the efficiency of corrections and complicate operational planning. Reliable short-term forecasts of atmospheric seeing could help optimize telescope scheduling, observation strategies, adaptive optics operation, and optical communication resource management.

Machine learning (ML) techniques have emerged as promising tools for atmospheric forecasting because they can model complex temporal relationships while providing rapid predictions. Previous studies have demonstrated the potential of data-driven approaches for seeing prediction. Short-term seeing forecasting has been explored using autoregressive integrated moving average (ARIMA) models~\cite{Kornilov_2015} as well as machine-learning techniques, including Random Forests and neural networks, for lead times of up to a few hours~\cite{Giordano_ML,Masciadri_1,Masciadri_2,Milli}. Longer-term turbulence forecasting using machine learning has also been investigated by~\citenum{Cherubini_ML}, while~\citenum{2022data} performed a comparative study of several data-driven methods for atmospheric turbulence prediction. Collectively, these studies demonstrate that machine learning can provide useful seeing forecasts, with most efforts focusing primarily on improving deterministic prediction accuracy.

However, atmospheric turbulence is inherently stochastic, and operational decisions often require information about forecast confidence in addition to point predictions. Probabilistic forecasting addresses this need by estimating a full predictive distribution rather than a single expected value. Such information is particularly valuable for risk-aware scheduling and decision-making, where understanding forecast uncertainty can be as important as minimizing prediction error. Among the available probabilistic machine learning approaches, Gaussian Processes (GPs) provide a well-established statistical framework that naturally produces predictive uncertainty estimates. More recently, normalizing-flow-based models have emerged as powerful probabilistic methods capable of learning complex, non-Gaussian predictive distributions directly from data.

In this work, we benchmark several machine learning approaches for short-term atmospheric seeing forecasting using historical seeing measurements from the Maunakea summit. We compare deterministic recurrent neural network (RNN) and long short-term memory (LSTM) models with two probabilistic approaches: Gaussian Processes (GPs) and a normalizing-flow-based model for time series (\textsc{FloTS}). Particular emphasis is placed on the comparison of the probabilistic models and their ability to provide reliable uncertainty quantification. Model performance is evaluated using both deterministic forecasting metrics and probabilistic calibration diagnostics. The objective of this study is to identify the most suitable framework for short-term seeing prediction and to assess the benefits of probabilistic forecasting for future operational applications in astronomy and optical communications.

\section{Dataset}
The dataset used in this study consists of atmospheric seeing measurements provided by the Maunakea Weather Center (MKWC), derived from a Monitoring Device (MD) combining the Multi-Aperture Scintillation Sensor (MASS) and the Differential Image Motion Monitor (DIMM)~\cite{Lyman}. The observations were collected at the summit of Maunakea, a high-altitude astronomical site known for its exceptional observing conditions, and span the period from September 2009 to November 2024. In this work, we focus on DIMM-derived seeing measurements, which represent the integrated optical turbulence across the full atmospheric column and provide estimates of total atmospheric seeing.
The seeing data range on the Maunakea summit varies between 0.04$^{\prime\prime}$ and 5$^{\prime\prime}$.

The raw time series, originally sampled at 90-second intervals, was preprocessed to ensure compatibility with machine learning models requiring regularly sampled inputs. This includes segmentation of continuous observing periods, handling of missing data, and resampling onto a uniform 10-minute grid using local linear interpolation. The resulting dataset preserves the temporal evolution of atmospheric conditions while reducing high-frequency variability that is less relevant for forecasting at operational timescales.

\section{Methods}
Atmospheric seeing is driven by turbulent fluctuations in the atmosphere, resulting in a stochastic and time-dependent process that is challenging to predict. Accurate forecasting therefore requires models capable of capturing temporal dependencies while, ideally, providing uncertainty estimates associated with their predictions.

Three forecasting approaches were investigated. First, Recurrent Neural Networks (RNNs) and Long Short-Term Memory (LSTM) networks were used as deterministic deep-learning baselines. Second, Gaussian Processes (GPs) were considered as a probabilistic statistical framework capable of providing uncertainty estimates. Finally, we evaluated \textsc{FloTS}, a probabilistic deep-learning model based on conditional normalizing flows that combines uncertainty quantification with the flexibility of deep neural networks.

All models were trained using a 2-hour history of seeing measurements to predict the subsequent 2 hours\cite{Milli,Turchi_2022,Masciadri_2}. This configuration was selected based on preliminary experiments showing that longer input windows did not improve forecasting performance.

\subsection{Recurrent Neural Network (RNN)/Long Short-Term Memory (LSTM)}
RNNs are neural networks specifically designed for sequential data\cite{RNN}. They process observations recursively, allowing information from previous time steps to influence future predictions through an internal hidden state. In this work, an encoder--decoder architecture was adopted, where the encoder compresses the historical seeing sequence into a latent representation, and the decoder generates future seeing values autoregressively.

For a standard RNN, the hidden state evolves according to

\begin{equation}
h_t=\sigma(W_hh_{t-1}+W_xx_t+b),
\end{equation}

where $x_t$ is the input at time step $t$, $h_t$ is the hidden state, and $\sigma$ denotes a nonlinear activation function.

Although RNNs are effective for modeling short-term temporal dependencies, they often struggle to retain information over long sequences due to the vanishing-gradient problem. To address this limitation, we also investigated Long Short-Term Memory networks\cite{lstm_1997,LSTM_book}. LSTMs extend the recurrent architecture through memory cells and gating mechanisms that regulate the flow of information, allowing relevant information to be retained over longer time scales.

The LSTM model follows the same encoder--decoder structure as the RNN but replaces recurrent units with LSTM cells. Both architectures consist of an encoder layer followed by two decoder layers with 100 hidden units per layer. A dropout rate of 0.3 was applied between layers to reduce overfitting.

Training was performed using the Adam optimizer with an initial learning rate of $10^{-4}$ and a Mean Squared Error (MSE) loss function,

\begin{equation}
\mathcal{L}_{\mathrm{MSE}}=
\frac{1}{N}
\sum_{i=1}^{N}
(y_i-\hat{y}_i)^2,
\end{equation}
where $y_i$ and $\hat{y}_i$ denote the observed and predicted seeing values, respectively.

While RNNs and LSTMs are capable of learning complex temporal relationships, they remain deterministic models and therefore provide only point forecasts without any estimate of predictive uncertainty.

\subsection{Gaussian Process (GP)}

To quantify forecast uncertainty, we investigated Gaussian Processes, a probabilistic framework that defines a distribution over functions\cite{GP_book}. Unlike deterministic neural networks, GPs naturally provide both a predictive mean and an associated confidence interval.

A GP is defined by a mean function $\mu(t)$ and a covariance function (kernel) $k(t,t')$,

\begin{equation}
f(t)\sim\mathcal{GP}\left(\mu(t),k(t,t')\right).
\end{equation}

The covariance kernel specifies the correlation between observations at different time steps and therefore controls the temporal structure learned by the model. In this work, we adopted an exponential kernel,

\begin{equation}
k(t,t')=
\sigma^2
\exp\left(-\frac{|t-t'|}{l}\right),
\end{equation}
where $\sigma^2$ is the signal variance and $l$ is the characteristic correlation length.

Given a sequence of past seeing measurements, the GP computes a posterior predictive distribution for future values. The predictive mean provides the forecast itself, while the predictive variance quantifies the associated uncertainty. This probabilistic formulation makes GPs particularly attractive for operational applications where confidence estimates are required.

Because GP training involves matrix inversions whose computational cost scales cubically with the number of training samples, a random subset of 500 training pairs was used during hyperparameter estimation. The model learned the kernel length scale, signal variance, and noise variance, which were subsequently averaged across training batches.

Although GPs provide an interpretable probabilistic framework, they assume Gaussian predictive distributions and can become computationally expensive for large datasets. Consequently, they are used here as a probabilistic baseline against which more flexible deep-learning approaches can be compared.

\subsection{Normalizing Flows (\textsc{FloTS})}

To overcome some of the limitations of GPs while retaining probabilistic forecasting capabilities, we employ \textsc{FloTS}, a forecasting framework based on conditional normalizing flows\cite{flow_1}.

Normalizing flows are generative models that transform a simple probability distribution into a more complex target distribution through a sequence of invertible mappings. Let $\mathbf{y}$ denote a seeing sequence and $\mathbf{z}$ a latent variable sampled from a Gaussian distribution. A normalizing flow learns an invertible transformation

\begin{equation}
\mathbf{y}=f_{\theta}(\mathbf{z}),
\end{equation}
where $\theta$ represents the trainable parameters of the model.

Using the change-of-variables theorem, the probability density of the target distribution can be written as

\begin{equation}
p_{\theta}(\mathbf{y})=
p(\mathbf{z})
\left|
\det
\frac{\partial f^{-1}_{\theta}}
{\partial \mathbf{y}}
\right|.
\end{equation}

The model parameters are optimized by maximizing the likelihood of the training data, allowing the flow to learn complex and potentially non-Gaussian distributions directly from observations.

In the forecasting setting, the objective is to model the conditional distribution of future seeing values given past measurements,

\begin{equation}
p(\mathbf{y}*{\mathrm{future}}
\mid
\mathbf{y}*{\mathrm{past}}).
\end{equation}
This enables the model to generate both point predictions and full predictive probability distributions.

The implementation adopted in this work is based on a Masked Autoregressive Flow (MAF)\cite{MAF}. MAF decomposes the target distribution into a sequence of autoregressive conditional probabilities, allowing efficient density estimation and sampling. To incorporate temporal information, the flow is conditioned on a context encoder that processes the historical seeing sequence.

Several encoder architectures were explored, including linear models, fully connected networks, RNNs, and Gaussian Processes. Consistent with previous studies, the best performance was obtained using an LSTM encoder, which effectively captures temporal dependencies in the seeing data while providing informative context to the flow.

Compared with GPs, \textsc{FloTS} offers two important advantages. First, it can represent highly flexible, non-Gaussian predictive distributions that may better reflect the statistical properties of atmospheric turbulence. Second, its computational scaling is more favorable for larger datasets. As a result, \textsc{FloTS} combines the uncertainty quantification capabilities of probabilistic forecasting with the expressive power of modern deep-learning architectures.


\section{Results}
\subsection{Forecasting performance}
\begin{figure} [ht]
   \begin{center}
   \begin{tabular}{c} 
   \includegraphics[width=\linewidth]{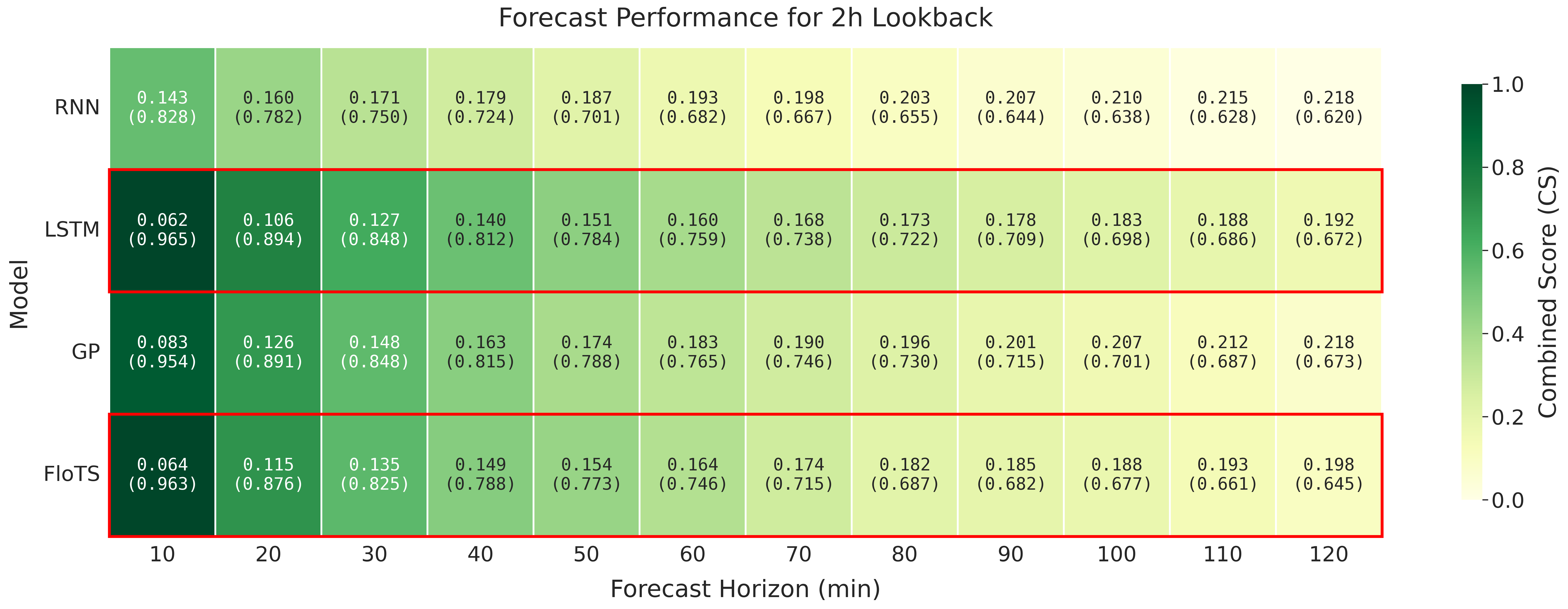}
	\end{tabular}
	\end{center}
   \caption[example] 
   { \label{fig:combined_results} 
Heatmap of combined forecast performance for each model over a 2-hour lookback window across forecast horizons. Each cell reports RMSE (top) and correlation (bottom, in parentheses). The background colour represents a composite performance score computed as the equally weighted average of min–max normalised RMSE (inverted) and correlation, where a score of 1 indicates the best performance and 0 the worst across all models and time steps. Red borders highlight the two best overall models, LSTM and FloTS, which consistently achieve the highest composite scores across all forecast horizons.}
   \end{figure}

Figure~\ref{fig:combined_results} presents the combined forecast performance of all models trained with a 2-hour lookback window across forecast horizons from 10 to 120 minutes. Each cell reports the Root Mean Square Error (RMSE, top) and Pearson correlation coefficient (bottom), while the background colour represents a composite performance score. This score is computed by first applying min-max normalisation independently to RMSE and correlation across all models and forecast horizons, scaling both metrics to the range $[0,1]$. Since lower RMSE indicates better performance whereas higher correlation is preferable, the normalised RMSE is inverted. The final combined score (CS) is then obtained as the equally weighted average of the inverted RMSE and the normalised correlation:

\begin{equation}
CS = \frac{1}{2}(1 - \mathrm{RMSE}_{\mathrm{norm}}) + \frac{1}{2}\mathrm{Corr}_{\mathrm{norm}}
\end{equation}
where $CS=1$ corresponds to the best joint performance (lowest RMSE and highest correlation), and $CS=0$ to the worst. This formulation allows both absolute prediction accuracy and temporal consistency to be evaluated simultaneously.

Across all models, performance degrades as the forecast horizon increases, with RMSE increasing and correlation decreasing, reflecting the growing difficulty of long-term seeing prediction. However, clear differences emerge between models. The LSTM consistently achieves the strongest combined performance across nearly all forecast horizons, maintaining the lowest RMSE and highest correlation, particularly in the short-term regime (10--60 min). \textsc{FloTS} follows closely behind, with competitive performance and only slightly higher errors than LSTM at most lead times. The GP model performs well in the short term, but its relative performance decreases more rapidly for longer horizons. In contrast, the standard RNN consistently yields the weakest combined scores, showing both larger prediction errors and weaker temporal agreement with observations.

The highlighted red rows in Figure~\ref{fig:combined_results} emphasise that LSTM and \textsc{FloTS} provide the most robust trade-off between accuracy and temporal consistency. While LSTM achieves the best deterministic forecasting performance overall, \textsc{FloTS} remains highly competitive while offering the additional advantage of flexible probabilistic forecasting (Section~\ref{sec:flots_results}).
   
\subsection{Uncertainty quantification}
\label{sec:flots_results}
Beyond point-forecast accuracy, probabilistic forecasting provides an estimate of the uncertainty associated with the future evolution of the seeing. Unlike the deterministic RNN and LSTM models, both GP and \textsc{FloTS} predict a full conditional probability density over the forecast horizon, allowing uncertainty to be quantified directly. However, the quality of this uncertainty estimate depends on calibration, i.e., the agreement between the predicted probability distribution and the empirical distribution of the observations. In this section, we focus on \textsc{FloTS} to illustrate how calibration affects the predictive density.

To improve this agreement, we apply a post-hoc temperature-based calibration to the \textsc{FloTS} predictive density \cite{Guo+2017,Srinivasan+25_LISA_GBs}. The calibration is based on the mismatch between the nominal coverage of the predicted credible regions and their empirical coverage measured on validation data. Following a temperature-scaling approach, the calibrated density is obtained as

\begin{equation}
p^*(x) = p(x)^{1/T^*(a)},
\end{equation}
where $p^*(x)$ and $p(x)$ denote the calibrated and uncalibrated predictive densities, respectively, and $T^*(a)$ is the optimal calibration temperature as a function of the nominal coverage $a$. In our case, $T^*(a)$ is parameterized as a bi-quadratic polynomial, whose coefficients are optimized to minimize the deviation between the empirical and nominal coverage in the probability-probability (PP) plot. Values of $T^*(a)>1$ broaden the predictive density to correct overconfidence, while $T^*(a)<1$ contracts it to correct underconfidence.

\begin{figure} [ht]
    \centering
    \includegraphics[width=0.9\linewidth]{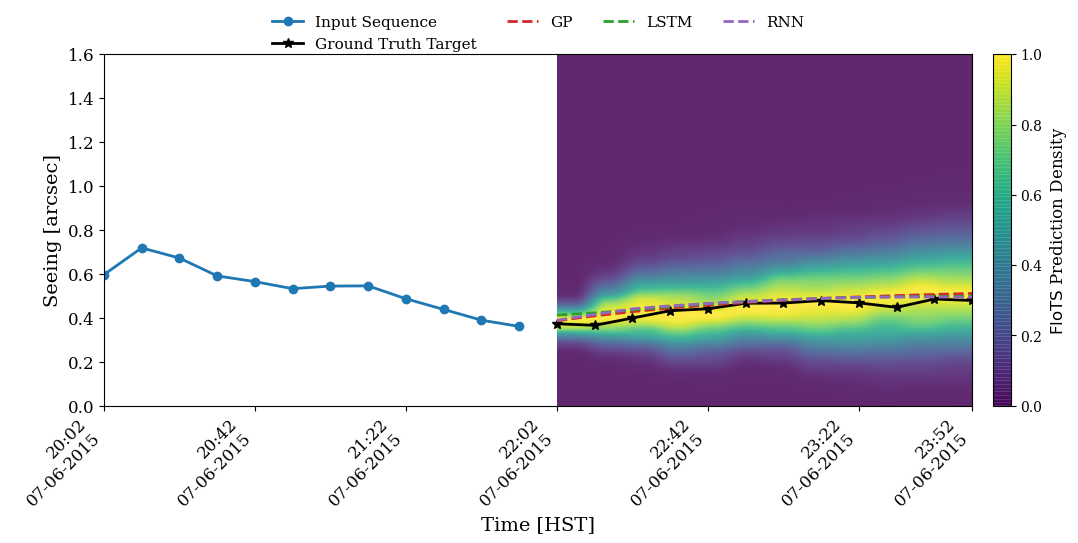}\\
    \includegraphics[width=0.9\linewidth]{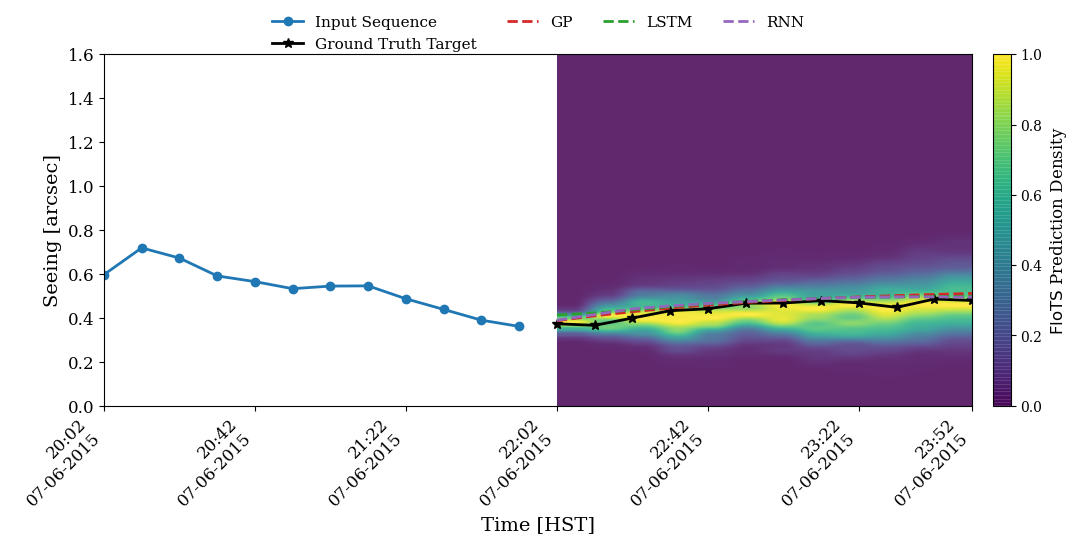}
    \caption{
    Comparison of the \textsc{FloTS} seeing prediction before and after calibration for the August case study.
    The top panel shows the uncalibrated prediction, while the bottom panel shows the calibrated prediction. In each panel, the left side shows the input sequence used for forecasting, and the right side displays the two-hour forecast horizon. The black curve indicates the observed target sequence, while the red, purple, and green dashed curves show the GP, RNN, and LSTM predictions, respectively. The heatmap represents the probability density predicted by \textsc{FloTS}. For visualization, the density is normalized independently at each forecast timestep. Calibration redistributes probability mass within the predictive density, modifying both the dispersion and the local non-Gaussian structure of the forecast distribution.
    }
    \label{fig:flow_calib}
\end{figure}

Figure~\ref{fig:flow_calib} illustrates an example forecast before and after calibration. In each panel, the left side shows the input seeing sequence used for conditioning, while the right side shows the 2-hour forecast horizon. The black curve represents the observed target sequence, and the dashed curves show the deterministic predictions from GP, RNN, and LSTM for comparison. The colored density map represents the \textsc{FloTS} predictive distribution of FloTS.

Before calibration (top panel), the predictive density is more widely spread around the central trajectory, reflecting an overdispersed and underconfident forecast. After calibration (bottom panel), the shape of the predictive distribution is adjusted by redistributing probability mass rather than shifting its central tendency. This concentrates likelihood around the observed seeing trajectory, placing the ground truth in regions of higher predicted probability and yielding a sharper, better-calibrated forecast. Unlike Gaussian-based approaches, where calibration mainly rescales the predictive variance, calibration of \textsc{FloTS} modifies the full predictive density and can preserve non-Gaussian structures such as asymmetry or multi-modality. This richer uncertainty representation makes \textsc{FloTS} particularly suitable for operational seeing forecasting, where understanding the range of plausible atmospheric conditions is as important as predicting the mean behavior.

\section{Conclusions}
In this work, we benchmarked several machine learning approaches for short-term atmospheric seeing forecasting using historical measurements from the Maunakea summit. Deterministic models (RNN and LSTM) and probabilistic models (GPs and \textsc{FloTS}) were evaluated over forecast horizons up to two hours using both accuracy-based and probabilistic metrics. The results show that forecasting performance degrades with increasing lead time for all models, as expected, but clear differences emerge in predictive capability. Among the deterministic approaches, the LSTM achieved the best overall point-forecast accuracy, consistently yielding the lowest RMSE and highest correlation across most forecast horizons.

Beyond deterministic performance, probabilistic forecasting provides valuable information on forecast confidence, which is essential for operational decision-making. While Gaussian Processes offer a natural probabilistic baseline, their Gaussian assumption limits the flexibility of the predictive distribution. In contrast, \textsc{FloTS} achieved competitive deterministic performance while producing richer, non-Gaussian predictive densities and more informative uncertainty estimates. The application of post-hoc temperature calibration further improved the reliability of these probabilistic forecasts by aligning predicted and empirical coverage.

These results highlight the potential of normalizing-flow-based models as a powerful framework for atmospheric seeing prediction, combining strong predictive accuracy with flexible uncertainty quantification. Future work will focus on extending the framework by incorporating additional meteorological and turbulence-related parameters, as well as exploring multi-site generalization for operational deployment in astronomy and free-space optical communication systems.


\acknowledgments 
The authors gratefully acknowledge the Maunakea Weather Center (MKWC), and in particular Dr. Steven Businger, Dr. Tiziana Cherubini, and Ryan Lyman, for providing the atmospheric data essential to this study. We also thank the Seeing Monitor Group for maintaining and making available the seeing measurements used in this work, as well as the staff of the Canada-France-Hawaii Telescope (CFHT) for their support in operating and maintaining the instruments and databases that enabled these observations. We further acknowledge that the summit of Mauna Kea is a place of profound cultural, spiritual, and scientific significance, and we recognize and respect its importance to the Native Hawaiian community.

RS acknowledges support from the European Union’s Horizon ERC Synergy Grant Making Sense of the Unexpected in the Gravitational-Wave Sky (Grant No. GWSky-101167314). Computational resources were partially provided through the SISSA–CINECA and CINECA–INFN agreements, which granted access to the LEONARDO supercomputing facility at CINECA. MB acknowledges financial support from the Centre National d’Études Spatiales (CNES) and Université Côte d’Azur (UCA). This work was also supported by the French Government through the UCAJEDI Investments in the Future program, managed by the National Research Agency (ANR) under Grant ANR-15-IDEX-01.
 
\bibliography{report} 
\bibliographystyle{spiebib} 

\end{document}